\documentclass[journal]{IEEEtran}
\usepackage{amsmath,amssymb}
\usepackage{psfrag}
\usepackage{epsfig}
\usepackage{cite}
\usepackage{graphics}
\usepackage{color}
\usepackage{subfigure}
\usepackage{multirow}
\usepackage{threeparttable}
\usepackage{booktabs}
\usepackage{mathrsfs}
\usepackage{bm}
\usepackage[ruled,vlined]{algorithm2e}

\usepackage{hyperref}
\usepackage{breakurl}

\begin{document}
\title{BAS: Beetle Antennae Search Algorithm for Optimization Problems}
\author{Xiangyuan Jiang, Shuai Li
%
}
\markboth{}%
{Shell \MakeLowercase{\textit{et al.}}: Bare Demo of IEEEtran.cls
for Journals} \maketitle
\begin{abstract}
Meta-heuristic algorithms have become very popular because of powerful performance on the optimization problem. A new algorithm called beetle antennae search algorithm (BAS) is proposed in the paper inspired by the searching behavior of longhorn beetles. The BAS algorithm imitates the function of antennae and the random walking mechanism of beetles in nature, and then two main steps of detecting and searching are implemented. Finally, the algorithm is benchmarked on 2 well-known test functions, in which the numerical results validate the efficacy of the proposed BAS algorithm.
\end{abstract}


\section{Introduction}
\IEEEPARstart{M}{eta-heuristic} algorithms have attracted an amount of attention over the last decades because of their simplicity, flexibility and local optimum avoidance. Specially, some of them play an fairly important role in not only academic society mainly concluding computer science but also many other practical engineering fields.
Mirjalili \emph{et al.} \cite{greywolf} propose a grey wolf optimizer by mimicking the leadership and prey of grey wolves in nature. Inspired by some cuckoo species' brood parasitism, Yang and Deb \cite{csvlf} develop a meta-heuristic algorithm called cuckoo search which could be applied to optimization and optimal search. Dorigo \emph{et al.} \cite{acsal} propose an ant colony optimization, which is inspired by the foraging movement of some ant species, to exploit for solutions of optimization problems. Two essential common characteristics of these different meta-heuristics are exploration and exploitation. This paper design a new meta-heuristic algorithm, which is named as beetle antennae search algorithm (BAS) to solve the optimization problems, taking inspiration from detecting and searching behavior of longhorn beetles.

\section{Beetle Antennae Search Algorithm Design}\label{sec.preliminary}

\begin{figure}[t]\centering
\subfigure[]{\includegraphics[scale=0.18]{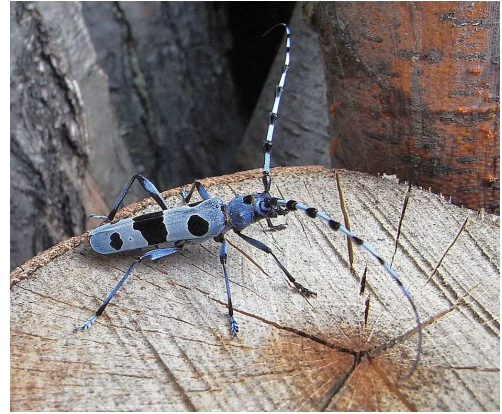}}
\subfigure[]{\includegraphics[scale=0.18]{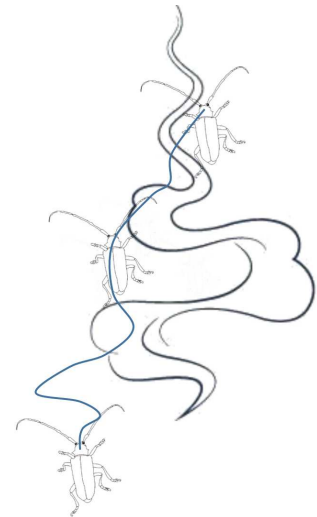}}
\caption{Longhorn beetle and its searching behavior with long antennae. (a) Longhorn beetle by Siga, used under the Creative Commons Attribution 3.0 Unported license \cite{beetle}. (b) Searching behavior of longhorn beetle with long antennae \cite{longhornbook}, where the line in black denotes the propagation of odour and the line in blue represents the trajectory of the beetle.}\label{fig.beetle}
\end{figure}

Longhorn beetles shown in Fig.\ref{fig.beetle}(a) are a family of beetles, characterized by extremely long antennae, which are often as long as or even longer than the beetle's body. The family is large with over 26,000 species described. Slightly most of them having long antennae. The pattern of antennae, which usually consists of many kinds of olfactory receptor cells, is often unique in particular species, and the functions of such sensing systems are still debating. However, two fundamental functions of such large antennaes are to bind to odours of prey, and to obtain the sex pheromone of potential suitable mate, in which large antennae could enlarge the detecting area. In addition, large antennae may also act as a protective warning mechanism.

We know that a beetle wobbles each antennae in one side of the body to receive the odour when it is preying or finding mates. That is to say, the beetle explores nearby area randomly using both antennae. Further more, when the antennae in one side detects a higher concentration of odour, the beetle would turn to the direction towards the same side, otherwise, it would turn to the other side. These two combined factors make most beetles to prey or seek for a mate as demonstrated in Fig.\ref{fig.beetle}(b), which inspire us to design a meta-heuristic optimization algorithm. Based on these two aspects, the basic steps of the proposed BAS algorithm can be summarized as the pseudo code shown in Algorithm \ref{alg.optim}.

Searching behavior of beetles with two antennae could be formulated in such a way that it is associated with an objective function to be optimized, which makes it possible to formulate new optimization algorithms. In the following sections, we will first outline the formulation of the BAS algorithm and then discuss the implementation as well as the analysis in detail.

For the sake of illustration, we denote the position of the beetle as a vector $\bm{x}^t$ at $t$th time instant ($t=1,2,\cdots$) and denote the concentration of odour at position $\bm{x}$ to be $f(\bm{x})$ known as a fitness function, where a maximum value of $f(\bm{x})$ corresponds to the source point of the odour.

For simplicity in describing our BAS algorithm, we now use the following two rules inspired by the behavior of beetle searching with antennae, which includes searching behavior and detecting behavior. Note that the beetle searches randomly to explore an unknown environment.

Firstly, to model the searching behavior, we propose describing a random direction of beetle searching as follows,
\begin{equation}\label{bearing}
  \overrightarrow{\bm{b}}=\frac{\text{rnd}(k,1)}{\|\text{rnd}(k,1)\|},
\end{equation}
where $\text{rnd}(.)$ denotes a random function, and $k$ represents the dimensions of position. Further more, we present the searching behaviors of both right-hand and left-hand sides respectively to imitate the activities of the beetle's antennae:
\begin{eqnarray}\label{xrxl}
  \bm{x}_r&=&\bm{x}^t+d^t\overrightarrow{\bm{b}}, \nonumber\\
  \bm{x}_l&=&\bm{x}^t-d^t\overrightarrow{\bm{b}},
\end{eqnarray}
where $\bm{x}_r$ denotes a position lying in the searching area of right-hand side, and $\bm{x}_l$ denotes that of the left-hand side. $d$ is the sensing length of antennae corresponding to the exploit ability, which should be large enough to cover an appropriate searching area to be cable of jumping out of local minimum points at the beginning and then attenuate as time elapses.

Secondly, to formulate the behavior of detecting, we further generate iterative model as follows to associate with the odour detection by considering the searching behavior,
 \begin{equation}\label{xupdaterule}
   \bm{x}^t=\bm{x}^{t-1}+\delta^t\overrightarrow{\bm{b}}\text{sign}(f(\bm{x}_r)-f(\bm{x}_l)),
 \end{equation}
where $\delta$ is the step size of searching which accounts for the convergence speed following a decreasing function of $t$ instead of an increasing function or a constant. The initialization of $\delta$ should be equivalent to the searching area. $\text{sign}(.)$ represents a sign function.

In terms of searching parameters i.e., antennae length $d$ and step size $\delta$, examples of update rules are presented for the designer as follows,
\begin{eqnarray}
  d^t&=&0.95d^{t-1}+0.01,\label{d.update}\\
  \delta^t&=&0.95\delta^{t-1}.\label{delta.update}
\end{eqnarray}
It is worth pointing out that both of the parameters could be designated as constants if necessary.
 \begin{algorithm}\label{alg.optim}
\caption{BAS algorithm for global minimum searching}

\KwIn{Establish an objective function $f(\bm{x}^t)$, where variable $\bm{x}^t=[x_1,\cdots,x_i]^\text{T}$ , initialize the parameters $\bm{x}^0,d^0,\delta^0$.}
\KwOut{$\bm{x}_\text{bst}$, $f_\text{bst}$.}
\While{ ($t<T_\text{max}$) or (stop criterion)}{
Generate the direction vector unit $\overrightarrow{\bm{b}}$ according to (\ref{bearing})\;
Search in variable space with two kinds of antennae according to (\ref{xrxl})\;
Update the state variable $\bm{x}^t$ according to (\ref{xupdaterule})\;
\If{$f(\bm{x}^t)<f_\text{bst}$}
{$f_\text{bst}=f(\bm{x}^t)$, $\bm{x}_\text{bst}=\bm{x}^t$.}
Update sensing diameter $d$ and step size $\delta$ with decreasing functions (\ref{d.update}) and (\ref{delta.update}) respectively, which could be further studied by the designers.
}

\Return{$\bm{x}_\text{bst}$, $f_\text{bst}$.}
\end{algorithm}

\section{Benchmark Validation}\label{sec.control}

To verify the efficacy of the proposed BAS algorithm, we use various benchmarks \cite{PSA} to validate the new algorithm.

\begin{figure}[t]\centering
\psfrag{data1}[c][c][0.5]{$f_\text{bst}$}
\psfrag{data2}[c][c][0.5]{$f(\bm{x}^t)$}
\subfigure[]{\includegraphics[scale=0.28]{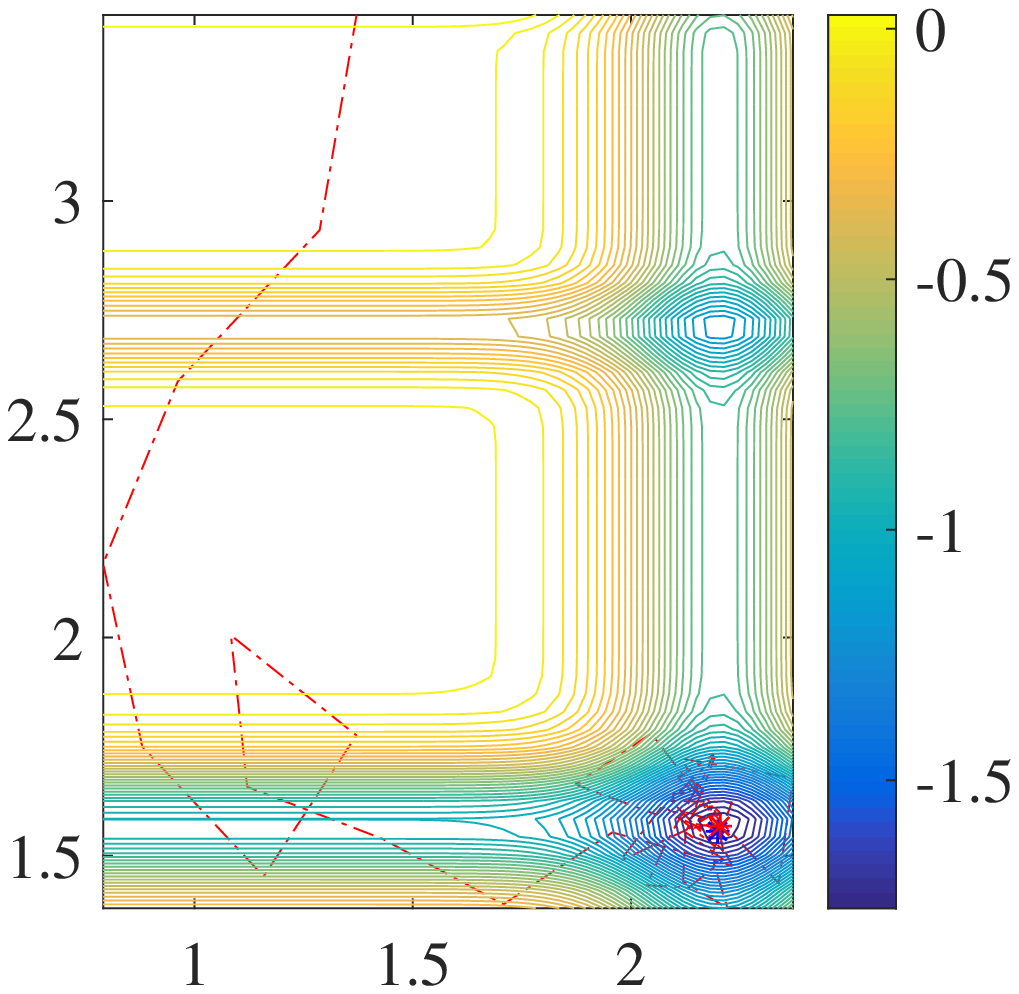}}
\subfigure[]{\includegraphics[scale=0.28]{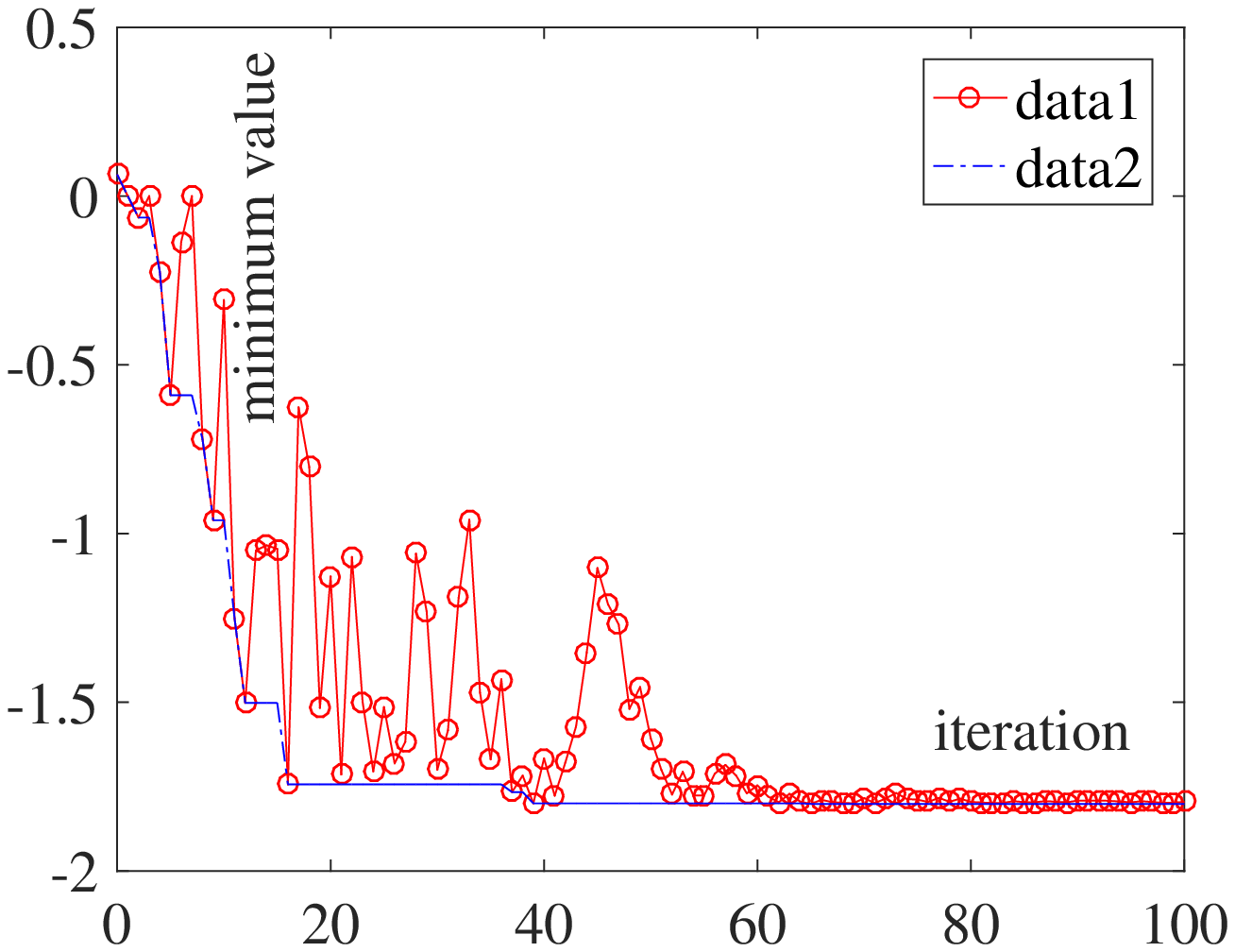}}
\caption{Performances of the proposed BAS algorithm to search the global optimum of Michalewicz function (\ref{fun.mich}) through $100$ iteration steps. (a) The searching trajectory of BAS algorithm when $\bm{x}$ lays in a 2-D space. (b) Convergence of the minimum value along iteration step $t$.}\label{fig.mich}
\end{figure}

At first, considering the Michalewicz function
\begin{equation}\label{fun.mich}
  f(\bm{x})=\sum_{i=1}^d\text{sin}(x_i)[\text{sin}(\frac{ix_i^2}{\pi})]^{2m},
\end{equation}
where $m=10$ and $i=1,2,\cdots$, the minimized value satisfies $f_*\approx-1.801$ locating in $\bm{x}_*\approx(2.20319, 1.57049)$ in $i=2$ dimension. The performance of BAS algorithm shows in Fig. \ref{fig.mich} along time step from $0$ to $100$ under the parameter configuration that sensing length $d$ updates following rule \ref{d.update} with initialized value $d^0=2$, and step size $\delta$ follows rule \ref{delta.update} with initialization $\delta^0=0.5$. It can be observed from Fig. \ref{fig.mich} that BAS algorithm could find the global minimum of function (\ref{fun.mich}). Numerically, the solution of function (\ref{fun.mich}) $f_\text{bst}=-1.8008$ is approximate to $f_*$ at the corresponding points $\bm{x}_{\text{bst}}=[2.1997,1.5679]^\text{T}$.

We also consider the Goldstein-Price function:
\begin{eqnarray}\label{fun.price}
  f(\bm{x})&=&[1+(x_1+x_2+1)^2(19-14x_1+3x_1^2-14x_2 \nonumber\\
            &&+6x_1x_2+3x_2^2)][30+(2x_1-3X_2)^2(18-32x_1\nonumber\\
            &&+12x_1^2+48x_2-36x_1x_2+27x_2^2)]
\end{eqnarray}
where the input domain is usually on the square $\{\bm{x} |x_i \in [-2,2], i=1,2 \}$.

\begin{figure}[t]\centering
\psfrag{data1}[c][c][0.5]{$f_\text{bst}$}
\psfrag{data2}[c][c][0.5]{$f(\bm{x}^t)$}
\subfigure[]{\includegraphics[scale=0.28]{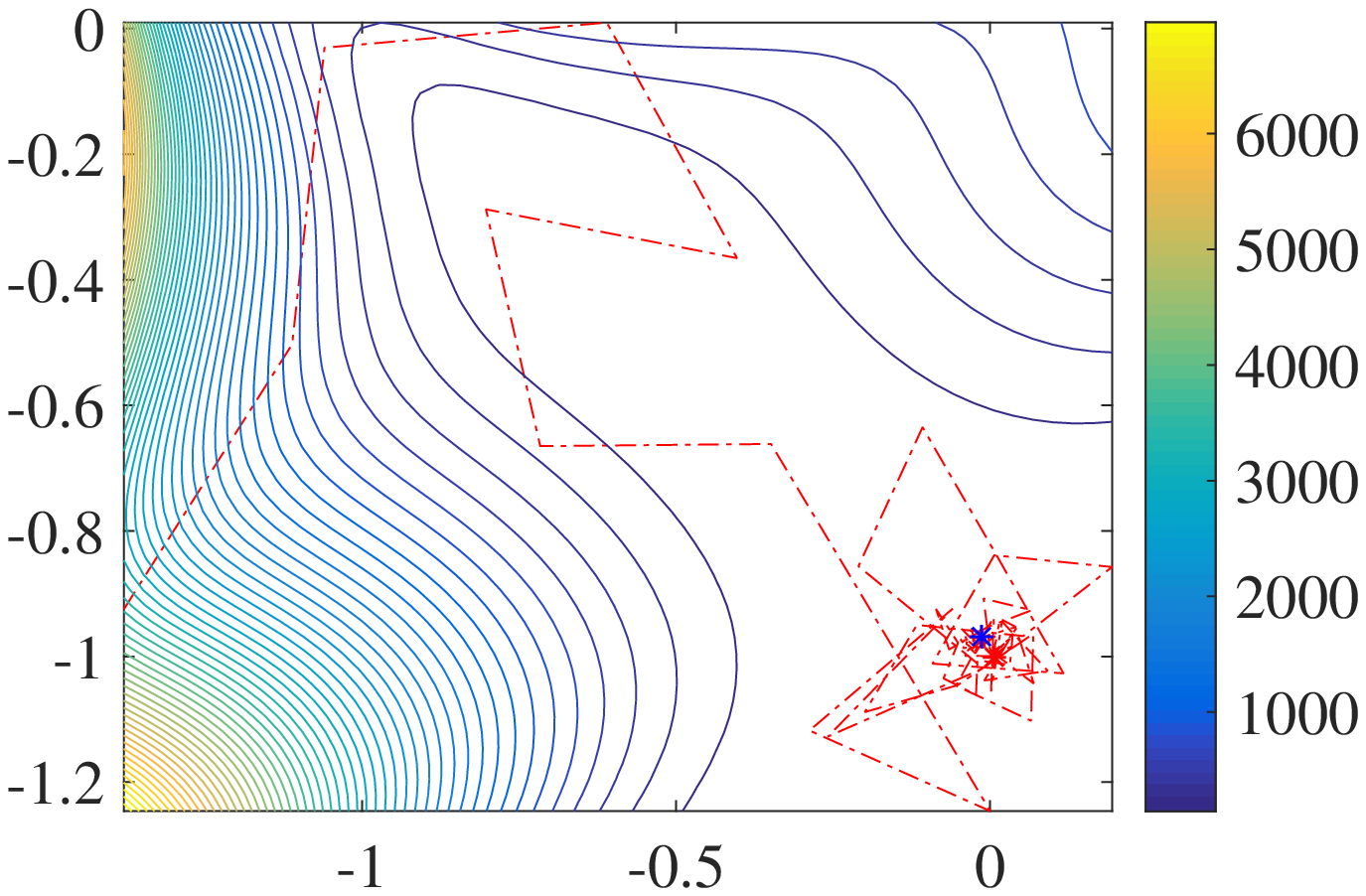}}
\subfigure[]{\includegraphics[scale=0.28]{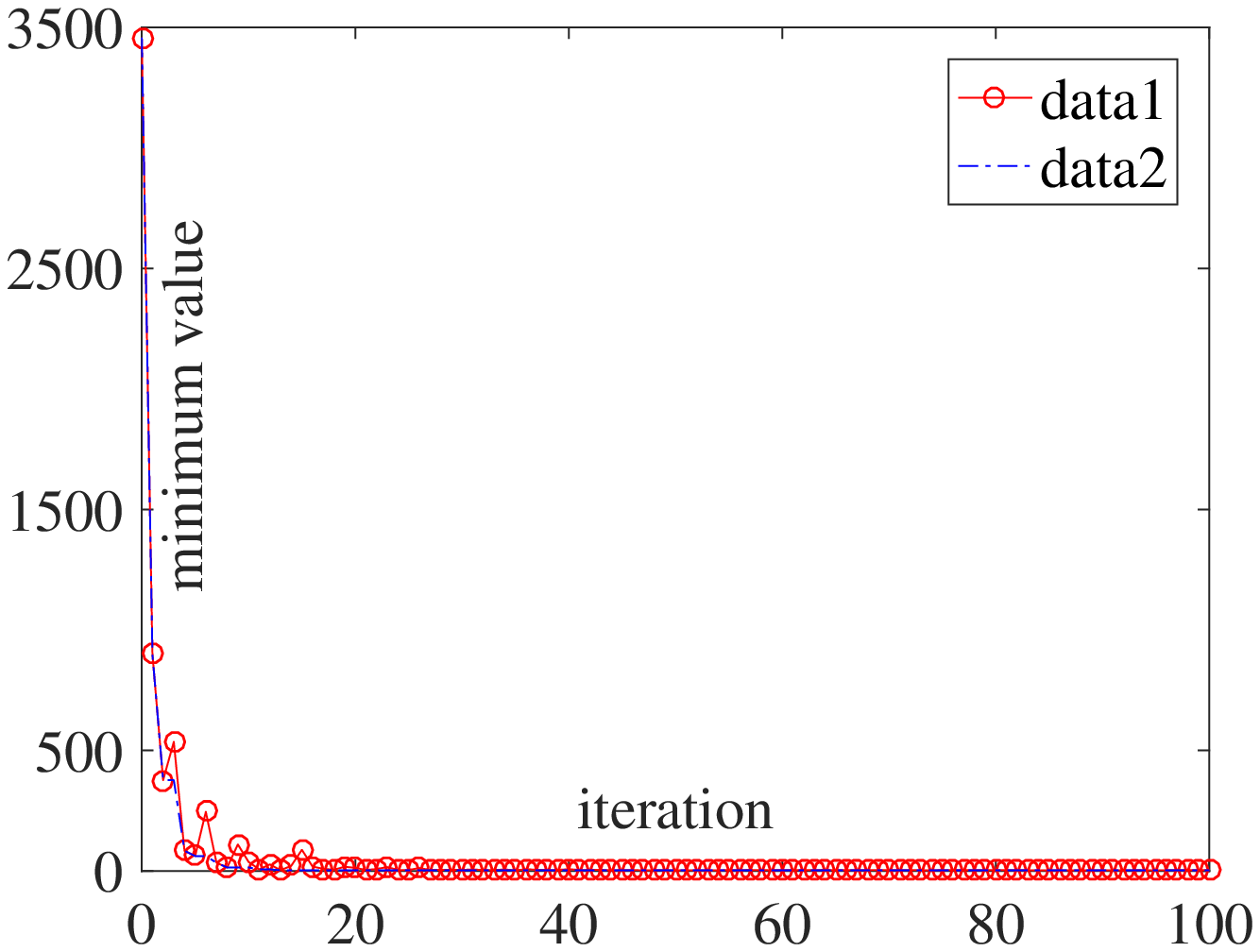}}
\caption{Performances of simulation results for the proposed BAS algorithm to solve the Goldstein and Price problem (\ref{fun.price}) through $100$ iteration steps. (a) The searching trajectory of BAS algorithm when $\bm{x}$ lays in a 2-D space. (b) Convergence of the minimum value along iteration step $t$.}\label{fig.price}
\end{figure}

There are several local minima for function (\ref{fun.price}) whose global minimum is $f(x^*)=3$ at $\bm{x}^*=[0,-1]^\text{T}$. By taking $100$ iterative steps and the parameters configurations the same as those in the aforementioned simulation, the visualization of simulation result of proposed BAS algorithm is demonstrated in Fig. \ref{fig.price}, in which (a) denotes the searching trajectory to seek the global optimum and (b) presents the convergence of the minimum along movement steps. The solution achieved in numerical experiment by the algorithm is $f_\text{bst}=3.0064$ at $\bm{x}_{\text{bst}}=[0.0052507,-0.99933]^\text{T}$ approaching $f_*$ at $x_*$. Results of the simulation substantiate the effectiveness of the proposed BAS algorithm sufficiently.

\section{Conclusion}\label{sec.conclusion}

This work presents a novel nature-inspired BAS algorithm to solve the optimization problem. This method mimics the detecting and searching behaviors of beetles and further parameter selections could be studied for the convergence of the algorithm. Two typical test functions are considered to benchmark performances of the algorithm in terms of convergence and local minimum avoidance. Both visualization results and numerical solutions validate the efficacy of the proposed algorithm.

\section{Appendix}

A numerical experiment of the proposed BAS algorithm in Matlab could be found at \url{https://www.mathworks.com/matlabcentral/fileexchange/64881-bas-beetle-antennae-search-algorithm-for-optimization}.

%
%
%
%
%

\end{document}